%% file: sample-acmtog-SIGGRAPH-submission.tex
\documentclass[acmtog]{acmart}
\acmSubmissionID{241}

\usepackage{booktabs} % For formal tables
\usepackage{bm}
\usepackage{multirow}
\usepackage{multicol}
\usepackage{color}
\definecolor{g}{RGB}{34,139,34}
\definecolor{b}{RGB}{30,144,255}
\usepackage{ulem}

\usepackage{color}
\definecolor{Red}{cmyk}{0,1,1,0}
\definecolor{Green}{cmyk}{1,0,1,0}
\definecolor{Cyan}{cmyk}{1,0,0,0}
\definecolor{Purple}{cmyk}{0.45,0.86,0,0}
\definecolor{Sky}{cmyk}{1.0,0.8,0,0}
\definecolor{Rosolic}{cmyk}{0.00,1.00,0.50,0}
\definecolor{Blue}{cmyk}{1.00,1.00,0.00,0}
\definecolor{Orange}{cmyk}{0,0.52,0.80,0}
\definecolor{Black}{cmyk}{1,0,0,1}

\newcommand{\yh}[1]{{\color{Black}{#1}}}

\newcommand{\ys}{\textcolor{Black}}
\newcommand{\secondary}{\textcolor{Black}}
\newcommand{\revise}{\textcolor{Black}}
\newcommand{\revisenew}{\textcolor{Black}}

\usepackage[ruled]{algorithm2e} % For algorithms

\SetAlFnt{\small}
\SetAlCapFnt{\small}
\SetAlCapNameFnt{\small}
\SetAlCapHSkip{0pt}

\acmJournal{TOG}
\begin{document}
% Title portion
\title{PVP-Recon: Progressive View Planning via Warping Consistency for Sparse-View Surface Reconstruction}

% DO NOT ENTER AUTHOR INFORMATION FOR ANONYMOUS TECHNICAL PAPER SUBMISSIONS TO SIGGRAPH 2019!
\author{Sheng Ye}
\email{yec22@mails.tsinghua.edu.cn}
\author{Yuze He}
\email{hyz22@mails.tsinghua.edu.cn}
\author{Matthieu Lin}
\email{yh-lin21@mails.tsinghua.edu.cn}
\author{Jenny Sheng}
\email{jsheng415@outlook.com}
\author{Ruoyu Fan}
\email{fry21@mails.tsinghua.edu.cn}
\affiliation{
 \institution{Tsinghua University}
 \city{Beijing}
 \country{China}
}

\author{Yiheng Han}
\affiliation{
 \institution{Beijing University of Technology}
 \city{Beijing}
 \country{China}
}
\email{hanyiheng@bjut.edu.cn}

\author{Yubin Hu}
\affiliation{
 \institution{Tsinghua University}
 \city{Beijing}
 \country{China}
}
\email{huyb20@mails.tsinghua.edu.cn}

\author{Ran Yi}
\affiliation{
 \institution{Shanghai Jiao Tong University}
 \city{Shanghai}
 \country{China}
}
\email{ranyi@sjtu.edu.cn}

\author{Yu-Hui Wen}
\authornotemark[1]
\affiliation{
 \institution{Beijing Jiaotong University}
 \city{Beijing}
 \country{China}
}
\email{yhwen1@bjtu.edu.cn}

\author{Yong-Jin Liu}
\authornote{Corresponding authors.}
\affiliation{
 \institution{Tsinghua University}
 \city{Beijing}
 \country{China}
}
\email{liuyongjin@tsinghua.edu.cn}

\author{Wenping Wang}
\affiliation{
 \institution{Texas A\&M University}
 \city{College Station}
 \country{USA}
}
\email{wenping@tamu.edu}

\begin{abstract}
Neural implicit representations have revolutionized dense multi-view surface reconstruction, yet their performance significantly diminishes with sparse input views. 
A few pioneering works have sought to tackle this challenge by leveraging additional geometric priors or multi-scene generalizability. However, they are still hindered by the imperfect choice of input views, using images under empirically determined viewpoints.
We propose \textit{PVP-Recon}, a novel and effective sparse-view surface reconstruction method that progressively plans the next best views to form an optimal set of sparse viewpoints for image capturing. \textit{PVP-Recon} starts initial surface reconstruction with as few as 3 views and progressively adds new views which are determined based on a novel warping score that reflects the information gain of each newly added view. This progressive view planning progress is interleaved with a neural SDF-based reconstruction module that utilizes multi-resolution hash features, enhanced by a progressive training scheme and a directional Hessian loss. 
Quantitative and qualitative experiments on three benchmark datasets show that our system achieves high-quality reconstruction with a constrained input budget and outperforms existing baselines.
\end{abstract}





\setcopyright{acmlicensed}
\acmJournal{TOG}
\acmYear{2024} \acmVolume{43} \acmNumber{6} \acmArticle{} \acmMonth{12}\acmDOI{10.1145/3687896}


\begin{teaserfigure}
  \setlength{\abovecaptionskip}{0pt}  
  \setlength{\belowcaptionskip}{0pt}
  \centering
  \includegraphics[width=0.95\textwidth]{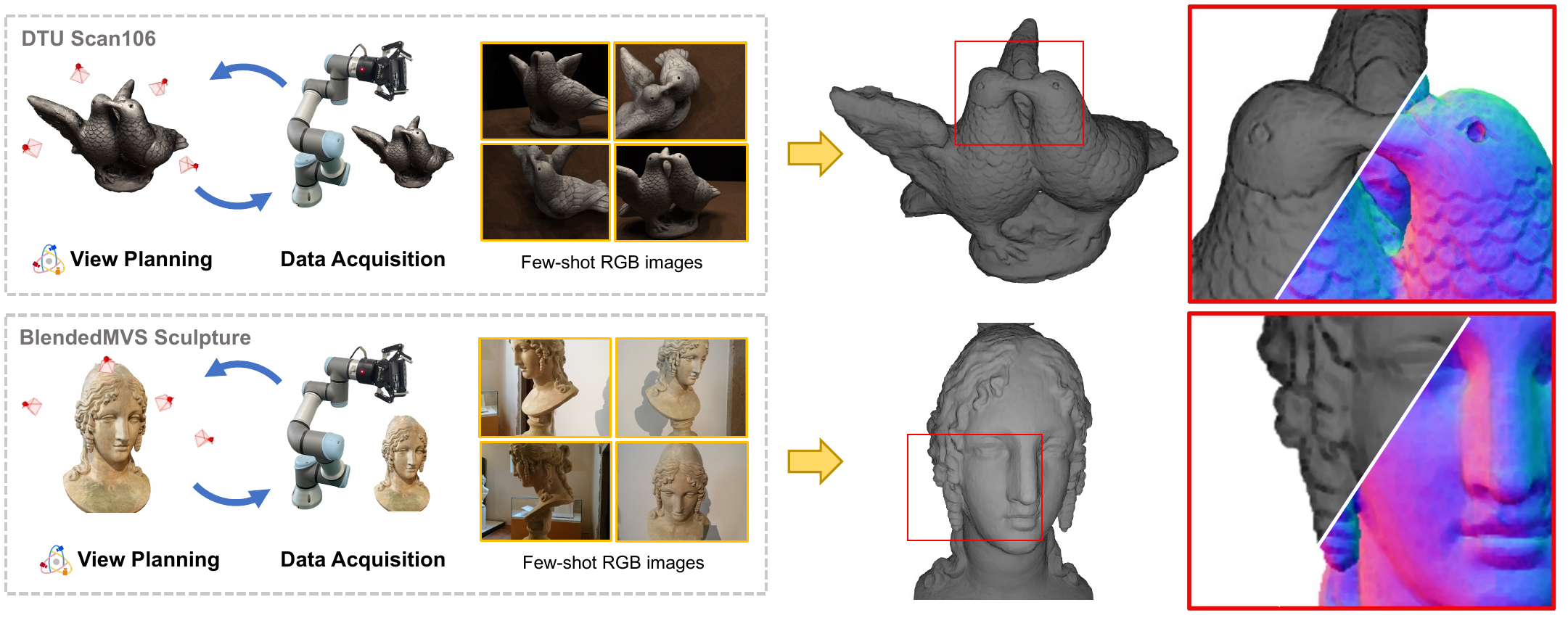}
  \caption{
  We present \textit{PVP-Recon}, a novel system that uses warping consistency to progressively determine the most informative views as input for surface reconstruction. \textit{PVP-Recon} produces high-quality mesh surfaces with a constrained budget on the number of input views. Colors indicate surface normals.
  }
  \label{fig:teaser}
  \Description{teaser}
  \vspace{0.3cm}
\end{teaserfigure}

\maketitle

\input{samplebody-journals}

\end{document}

%% file: samplebody-journals.tex
\section{Introduction}
3D mesh surface reconstruction from multi-view RGB images has always been an important issue in computer vision and graphics.
Classic multi-view stereo (MVS) algorithms~\cite{colmap, mvsnet} are well established in the field of 3D reconstruction, but struggle to handle areas with homogeneous textures.
Recently, neural implicit methods~\cite{neus, volsdf} propose to represent the scene as a signed distance field (SDF) and optimize mesh surfaces directly from 2D image supervision through differentiable rendering.
Although neural implicit methods surpass previous works in achieving high-fidelity mesh reconstruction, they typically rely on densely captured images as inputs, whose acquisition process can be time-consuming.
To alleviate this reliance, some methods attempt to reconstruct 3D mesh from a predefined set of sparse views, based on geometric priors~\cite{monosdf} or generalizable priors~\cite{sparseneus}.
Evidently, the surface reconstruction quality will depend on the input sparse views. \revise{However, it remains unknown how to determine a suitable set of sparse input views to achieve good surface reconstruction of a given object.}

In this paper, we observe that two key factors contribute to good surface reconstruction under sparse views: (1) an effective view planning strategy to provide the most informative input images; and (2) a suitable neural representation with well-designed regularization techniques to encode the geometry.
%Moreover, these two factors are closely related, as the planning strategy should fit the geometric representation.
Thus, we propose \textit{PVP-Recon}, a new sparse-view surface reconstruction %method 
\yh{system} that contains two modules: (1) a view planning module for identifying most informative viewpoints for additional image capturing on the fly during surface optimization; and (2) a neural surface reconstruction module utilizing supplemented additional views to gradually improve reconstruction quality. \secondary{The modules are easy to use and highly flexible, and they can be switched out in a modular manner against other existing approaches.}

Different from existing sparse-view reconstruction methods, our method does \textit{not} use a predefined set of sparse images as input. \yh{Instead, we use our view planning module to progressively identify the most informative viewpoints for capturing additional images during the reconstruction process until a high-quality surface is obtained.} 
Hence, we have essentially solved a \textit{next-best-view} problem~\cite{NBV} in the setting of image-based surface reconstruction.
We propose a novel warping-based strategy, which selects the viewpoint with maximum information gain for subsequent reconstruction by calculating the cross-view rendering consistency for each candidate viewpoint.
As a result, \textit{PVP-Recon} typically ends up successfully reconstructing an object using no more than 8 images.

Our reconstruction module uses multi-resolution hash features to represent the SDF of object surfaces due to their expressiveness. \yh{Moreover, we introduce a progressive training scheme that gradually activates higher-resolution hash features, and a novel directional Hessian loss \secondary{to regularize the neural SDF field and encourage the SDF under sparse views to be more smooth}, which in combination make the optimization process more robust by preventing the model from quickly overfitting. The reconstruction module can provide renderings to the view planning module for decision-making, and facilitates the optimization of hash features under progressively added sparse views with the directional Hessian loss.
} 
%, and provides renderings to the view planning module for decision-making.
%We note that the optimization of hash features under progressively added sparse views can be ill-posed, hindering the final reconstruction results and the effectiveness of view planning.
%Therefore, we introduce a progressive training scheme that gradually activates higher-resolution hash features, and a novel directional Hessian loss to regularize the neural SDF field, which in combination make the optimization process efficient and robust by preventing the model from quickly overfitting.

In summary, we make the following three contributions: (1) A novel object-level surface reconstruction system from sparse input views, where the inputs are progressively supplemented during the surface optimization process; (2) An effective view planning strategy based on image warping consistency; (3) A progressive training scheme enhanced by a directional Hessian loss to facilitate high-quality 3D reconstruction.% from sparse views.

\begin{figure*}[t]% specify a combination of t, b, p, or h for top, bottom, on its own page, or here
  \centering % avoid the use of \begin{center}...\end{center} and use \centering instead (more compact)
  \setlength{\abovecaptionskip}{4pt}  
  \setlength{\belowcaptionskip}{0pt}
  \includegraphics[width=1.0\linewidth]{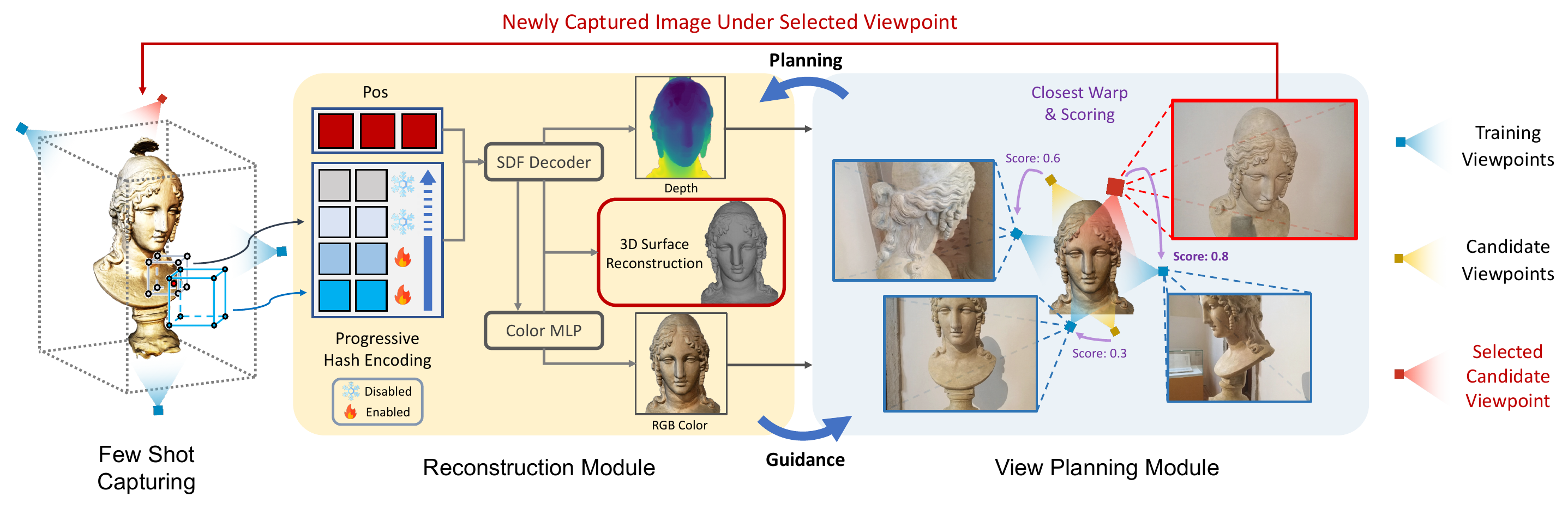}
  \caption{
    Overview of our proposed system. The overall framework consists of two collaborative modules. In the reconstruction module, we use hash encoding to represent the SDF of object geometry, and adopt a training scheme that progressively enables finer levels of hash features. In the view planning module, we design a simple yet effective warping-based scoring strategy that progressively supplements the most informative views for surface reconstruction.
  }
  \label{fig:pipeline}
  \Description{pipeline}
\end{figure*}

\section{Related Works}

\subsection{Neural Implicit Representation}
Recently, neural implicit representations have emerged as a powerful tool to encode the 3D geometry of a target object due to their compactness and remarkable performance.
Occ-Net~\cite{occ-net} and DeepSDF~\cite{deepsdf} first propose to use neural networks to model shapes as occupancy or signed distance functions. However, they require ground-truth 3D data to supervise the learning process.
Some follow-up works try to incorporate neural implicit functions and differentiable rendering to recover surfaces only with multi-view 2D image supervision.
Specifically, IDR~\cite{IDR} designs a differentiable rendering framework for implicit shape and appearance representations.
NeuS~\cite{neus} and VolSDF~\cite{volsdf} utilize signed distance fields to represent implicit surfaces and adopt the volume rendering technique introduced in NeRF~\cite{nerf}.
To enhance the expressive capability, Neuralangelo~\cite{neuralangelo} replaces the MLPs used in previous methods with multi-resolution hash grid features to encode implicit geometric functions.

\subsection{Sparse View Reconstruction}
Although neural implicit methods can produce remarkable meshes with dense input views, their performance drops drastically under sparse views due to the shape-radiance ambiguity~\cite{nerf++}.
To address this challenge, a few pioneering works have been proposed, which can be divided into two main categories: regularization-based and generalization-based.
Regularization-based methods utilize semantic priors~\cite{dietnerf}, geometric priors~\cite{ds-nerf, monosdf}, frequency priors~\cite{freenerf}, or MVS priors~\cite{s-volsdf} as extra constraints, facilitating the ill-posed optimization under sparse views.
Generalization-based methods aim to train on multiple scenes to gain generalization to unseen scenes. There have already been some successful attempts~\cite{mvsnerf, pixelnerf, geonerf} at neural rendering.
In terms of surface reconstruction, SparseNeuS~\cite{sparseneus} learns generalizable priors from image features and adopts cascaded volumes for surface prediction.
DiViNeT~\cite{DiViNeT} learns a set of templates across different scenes, which serve as anchors in new scenes.
To generate more details, VolRecon~\cite{volrecon} represents the geometry as the Signed Ray Distance Function (SRDF) and combines multi-scale features to regress SRDF values using a ray transformer.
However, neither regularization-based nor generalization-based methods consider the importance of input view planning for sparse reconstruction. Regularization-based methods typically fix a few images as input before training, while generalization-based methods only work well under cautiously selected images with large overlap~\cite{DiViNeT}.

\subsection{Neural Implicit View Planning}
Some recent methods~\cite{active-3d, activenerf, nerf-ensemble, neu-nbv, neurar} attempt to adopt neural implicit representations for robot active planning.
These methods determine the next best view based on uncertainty estimation with different policies. Most of these active planning methods~\cite{activenerf, nerf-ensemble, neu-nbv} focus on synthesizing novel views rather than 3D surface reconstruction.
Closely related to our work, two studies~\cite{active-3d, neurar} also aim at view planning for 3D surface reconstruction. \citet{active-3d} propose to calculate the density entropy along each ray as the uncertainty measurement, which is reasonable but susceptible to noise. NeurAR~\cite{neurar} models the emitted radiance as Gaussian distributions and evaluates the uncertainty by aggregating the variance. However, this approach may lead to degraded quality and training instability.
The aforementioned methods mainly concentrate on designing uncertainty estimation strategies of implicit representations, without considering discernible bias in surface reconstruction. We propose a simple yet effective view planning strategy for high-quality object-level surface reconstruction under sparse views by evaluating the multi-view consistency.

\section{Method}
\yh{We propose \textit{PVP-Recon}, a novel system consisting of a view planning module and a reconstruction module, to reconstruct high-quality object surfaces from a sparse set of RGB images. Specifically, the view planning module uses a warping-based strategy to determine the most informative viewpoint for subsequent image capture (Section~\ref{subsec:VSmodule}). The reconstruction module adopts a \textit{progressive hash encoding} as the geometric representation, facilitated by a novel directional Hessian loss, for reconstructing high-quality object surfaces (Section~\ref{subsec:ReconModule}). The overall framework is illustrated in Figure~\ref{fig:pipeline}. 

We assume that view planning and surface reconstruction are mutually reinforcing. The newly acquired image from the view planning module is supplemented to the reconstruction module to help its optimization of recovering 3D surfaces. In turn, the reconstruction module provides current optimization status to guide the view planning module to make further planning decisions.
We alternately plan subsequent input views and optimize the mesh surface, trying to achieve the best reconstruction results with a few images. 
}

% We propose \textit{PVP-Recon}, a novel framework that progressively finds and adds the best next views for reconstructing high-quality object surfaces from a sparse set of RGB images.
% The overall framework is illustrated in Figure~\ref{fig:pipeline}, consisting of a view planning module (Section~\ref{subsec:VSmodule}) and a reconstruction module (Section~\ref{subsec:ReconModule}).

% The key to our framework is the collaborative interaction between the two modules.
% We assume that surface reconstruction and view planning are mutually reinforcing.
% The reconstruction module utilizes few-shot RGB images dynamically maintained by the view planning module to recover 3D surfaces, and in turn provides current optimization status to guide the view planning module to make reasonable planning decisions.
% The view planning module uses a warping-based strategy to determine the most informative viewpoint for subsequent image capture. The newly acquired image is then supplemented to the reconstruction module to help its further optimization.
% We alternately optimize the mesh surface and plan subsequent input views, trying to achieve the best reconstruction results with a few images.
% To stabilize the optimization under progressively added sparse views, our reconstruction module adopts a \textit{progressive hash encoding} as the geometric representation, facilitated by a novel

% Below, we first briefly revisit some preliminaries (Section~\ref{subsec:pre}).

\subsection{Preliminaries}
\label{subsec:pre}
\subsubsection{Multi-resolution hash encoding}
Instant-NGP~\cite{instant-ngp} first proposes a multi-resolution hash encoding representation.
The hash encoding partitions the space into multi-resolution grids of $L$ levels $\{ V_1, ..., V_L \}$.
Using the spatial hash function~\cite{hash_func}, each grid cell corner is mapped to a hash entry that stores a learnable feature vector.
Given a 3D point $\bm{x}$, we map it to different locations $\bm{x}_l$ at each grid resolution $V_l$ and retrieve the corresponding features $\gamma_{l}(\bm{x}_l) \in \mathbb{R}^{F}$ via trilinear interpolating hash entries at cell corners.
The feature vectors of each level and auxiliary inputs $\xi \in \mathbb{R}^E$ (such as the view direction) are concatenated together to form the final encoded feature:
\begin{equation}
\label{eqn:hash_encoding}
\gamma(\bm{x}) = (\xi, \gamma_1(\bm{x}_1), ..., \gamma_L(\bm{x}_L)).
\end{equation}
The encoded feature $\gamma(\bm{x}) \in \mathbb{R}^{LF+E}$ is then passed to a shallow MLP decoder to regress the final output.
Hash collisions are not handled explicitly, as the decoder is assumed to disambiguate collisions.

\subsubsection{SDF-based volume rendering}
NeuS~\cite{neus} proposes to encode 3D scene as a signed distance field $f_g(\bm{x}; \theta_g)$ that outputs the SDF value of any specified location $\bm{x}$, and a color field $f_c(\bm{x}, \bm{d}; \theta_c)$ that outputs the view-dependent emitted radiance.
Both fields can learn from 2D image supervision using the differentiable volume rendering technique.
Given a ray $\bm{r}$ emitted from a posed camera, the rendering scheme first samples a set of points $\{ \bm{x}_1, \bm{x}_2, ..., \bm{x}_N \}$ along the ray. 
Then, the rendered pixel color $C(\bm{r})$ of this ray $\bm{r}$ can be calculated as:
\begin{equation}
\label{eqn:volume_render}
C(\bm{r}) = \sum_{i=1}^{N}{T_i\alpha_i f_c(\bm{x}_i,\bm{d})}, \ T_i = \prod_{j=1}^{i-1}{(1 - \alpha_j)},
\end{equation}
where $\bm{d}$ is the direction vector of ray $\bm{r}$, and $\alpha_i$ denotes the discrete opacity of the $i$-th ray segment. Formally, $\alpha_i$ is defined as:
\begin{equation}
\alpha_i=\max \left(0, \frac{\Phi_\tau(f_g(\bm{x}_i)) - \Phi_\tau(f_g(\bm{x}_{i+1}))}{\Phi_\tau(f_g(\bm{x}_i))} \right),
\end{equation}
where $\Phi_\tau$ is the Sigmoid function with a learnable parameter $\tau$. 
The final surface $S$ can be extracted as the zero level-set of the SDF, \textit{i.e.}, $S = \{ \bm{x} \in \mathbb{R}^3 | f_g(\bm{x}) = 0 \}$, by the Marching Cubes method.

\subsection{View Planning Module With Warping Consistency}
\label{subsec:VSmodule}
Our view planning module maintains a %sparse 
set of images for surface recovery, whose number dynamically increases as training proceeds.
This module first discretizes the continuous camera space into a set of candidate viewpoints. Here, viewpoints refer to camera poses from which subsequent images can be captured, and can be easily obtained through uniform sampling or predefined.
Then, we cluster all candidate camera poses and obtain the images under the cluster center poses as initial training images (typically 3 views).
At regular training intervals, we score each remaining candidate pose and supplement the training image set with a newly captured image under the highest-scoring pose, until the view number reaches a given threshold.
Note that when our view planning strategy evaluates the set of candidate camera poses, we assume that no images under these poses are already captured or available.

\begin{figure}[t]% specify a combination of t, b, p, or h for top, bottom, on its own page, or here
  \centering % avoid the use of \begin{center}...\end{center} and use \centering instead (more compact)
  \setlength{\abovecaptionskip}{4pt}  
  \setlength{\belowcaptionskip}{0pt}
  \includegraphics[width=\columnwidth]{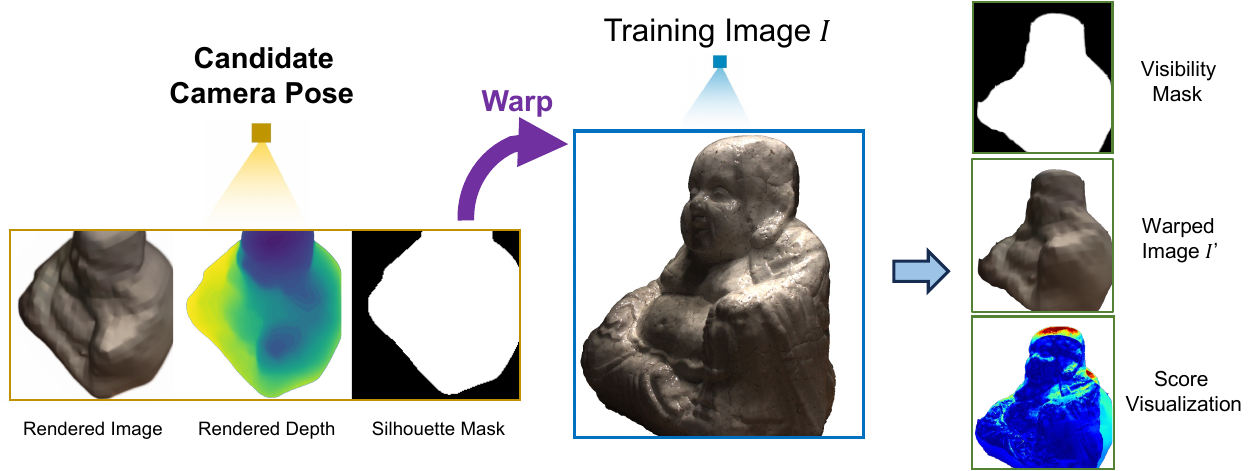}
  \caption{%
    Illustration of our view planning strategy \revise{on DTU scan114}. \revise{For each candidate view, we render the current reconstructed mesh from this view. Then, we use the rendered depth and silhouette mask to warp the rendered RGB image into the closest training view and evaluate the warping score.}
  }
  \label{fig:warp_score}
  \Description{warp_score}
\end{figure}

The main challenge is to calculate a reasonable score for each candidate camera pose that indicates its potential contribution (the amount of additional information it can provide) to future reconstruction.
Our key insight is that the consistency of rendered images and depths from the reconstruction module 
can effectively reflect whether a specific area needs more information from an additional image.
We aim to examine the multi-view consistency of renderings by drawing attention from the image-warping technique.
Specifically, denote the relative transformation matrix between two camera poses as $T$, the camera intrinsic matrix as $K$, the depth map as $d$, then the image-warping process can be computed as:
\begin{equation}
\label{eqn:image_warp}
p' = K \mathcal{F}^{-1} \{ T \mathcal{F} \{ d(p) K^{-1} p \} \}, 
\end{equation}
where $p$ and $p'$ are the homogeneous locations of the source image and the warped image. $\mathcal{F}$ is the homogeneous conversion from $3 \times 1$ to $4 \times 1$ vectors.
To calculate the warping score for a specific candidate pose, we first render the image, depth, and silhouette mask from our reconstruction module under that pose.
The rendered image is then warped to its closest training view, \textit{i.e.,} an existing training image whose pose is closest to this candidate pose. The silhouette mask is also warped and acts as a visibility mask, indicating which areas of this training view are visible to the rendered image.
The warping score $\bm{s}$ is defined as the \textit{photometric difference} between the warped image $I'$ and the training image $I$ within the visibility mask $M$:
\begin{equation}
\label{eqn:score}
\bm{s} = \left \| M \odot (I' - I) \right \|_1.
\end{equation}

Figure \ref{fig:warp_score} illustrates our proposed planning strategy. There are two main factors contributing to a high warping score as defined in Eqn. \ref{eqn:score}. First, this photometric difference between $I'$ and $I$ can be caused by incomplete training of the 3D neural representation, revealing errors in current reconstructed geometry. Second, the difference between $I'$ and $I$ can be attributed to the multi-view inconsistency caused by occlusion, that is, there are some regions of the current reconstructed 3D model that are visible to the view under candidate pose, but not present in the training image $I$, suggesting that a new view under this candidate pose is likely to provide coverage to new regions that have not been covered by the existing training views. While the definition of our warping score takes both factors into account, we observe that the latter contributes more to the score.
To sum up, by progressively selecting the highest-scoring viewpoints, we attempt to seek the best next views with maximum information gain to improve subsequent surface reconstruction.

\begin{small}
\begin{table*}[tb]
\centering
\setlength{\abovecaptionskip}{3pt}  
\setlength{\belowcaptionskip}{0pt}
\caption{Quantitative results on DTU dataset (bold means best, underline means second best). We report the Chamfer distance (mm) $\downarrow$.
}
\renewcommand\arraystretch{0.8}
\setlength{\tabcolsep}{4.25pt}
\begin{tabular}{c|c|ccccccccccc}
\toprule
\textbf{Methods} & \textbf{Time}& \textbf{scan55} & \textbf{scan65} & \textbf{scan69} & \textbf{scan83} & \textbf{scan105} & \textbf{scan106} & \textbf{scan110} & \textbf{scan114} & \textbf{scan118} & \textbf{scan122} & \textbf{mean} \\
\midrule
NeuS (\textit{random}) & \multirow{3}{*}{83min} & 1.44 & 2.03 & 1.21 & 1.66 & 1.01 & 0.99 & 3.60 & 0.48 & 1.14 & 1.23 & 1.48 \\
NeuS (\textit{cluster}) & & 0.76 & 1.90 & 1.17 & 1.57 & 0.98 & 0.98 & 3.11 & 0.49 & 0.96 & 0.77 & 1.27 \\
NeuS (\textit{farthest}) & & 1.45 & 1.76 & 1.87 & 1.62 & 1.25 & 1.26 & 3.35 & 0.63 & 1.37 & 1.04 & 1.56 \\
\midrule
Neuralangelo (\textit{random}) & \multirow{3}{*}{512min} & 0.54 & 1.18 & 1.36 & 1.62 & 0.96 & \uline{0.80} & 2.62 & 1.54 & \uline{0.79} & 0.91 & 1.23 \\
Neuralangelo (\textit{cluster}) & & \uline{0.52} & 1.43 & 1.15 & 1.43 & \textbf{0.74} & \textbf{0.65} & 2.98 & 0.99 & 0.88 & 0.70 & 1.15 \\
Neuralangelo (\textit{farthest}) & & 0.70 & 1.96 & 1.66 & 1.39 & 0.99 & 0.86 & 3.14 & 1.42 & 1.67 & 0.78 & 1.46 \\
\midrule
MonoSDF (\textit{random}) & \multirow{3}{*}{435min} & 0.79 & 1.24 & 1.19 & 1.44 & \uline{0.80} & 3.05 & 1.53 & 0.68 & 1.34 & 2.34 & 1.44 \\
MonoSDF (\textit{cluster}) & & 0.89 & \textbf{1.07} & 1.09 & \uline{1.35} & 0.97 & 2.74 & 1.29 & 0.71 & 1.38 & 1.73 & 1.32 \\
MonoSDF (\textit{farthest}) & & 0.98 & 1.53 & 1.39 & 1.37 & 0.91 & 2.64 & 1.90 & 0.86 & 1.68 & 1.45 & 1.47 \\
\midrule
SparseNeuS & $-$ & 0.84 & 1.87 & 1.07 & 1.51 & 1.26 & 1.11 & \uline{1.09} & 0.74 & 1.46 & 1.83 & 1.28 \\
\midrule
VolRecon & $-$ & 0.92 & 1.92 & 1.01 & 1.58 & 0.89 & 1.09 & 1.48 & 0.63 & 1.20 & 1.12 & 1.18 \\
\midrule
Ours (\textit{random}) & \multirow{4}{*}{9min} & 0.64 & 1.34 & 0.84 & 1.52 & 0.95 & 1.22 & 1.13 & 0.48 & 0.93 & 0.62 & 0.97 \\
Ours (\textit{cluster}) & & 0.55 & 1.30 & \uline{0.77} & 1.39 & 0.87 & 0.92 & 1.35 & \uline{0.47} & 0.80 & 0.55 & \uline{0.90} \\
Ours (\textit{farthest}) & & 0.76 & 1.38 & 1.22 & 1.43 & 0.89 & 1.05 & 1.47 & 0.49 & 0.91 & \uline{0.51} & 1.01 \\
Ours (\textit{planning}) & & \textbf{0.51} & \uline{1.15} & \textbf{0.75} & \textbf{1.28} & 0.84 & 0.91 & \textbf{0.95} & \textbf{0.46} & \textbf{0.72} & \textbf{0.50} & \textbf{0.81} \\
\bottomrule
\end{tabular}
\label{tab:dtu_comp}
\end{table*}
\end{small}

\subsection{Reconstruction Module With Progressive Scheme}
\label{subsec:ReconModule}
In this module, we represent the object geometry as a signed distance field (SDF) encoded by multi-resolution hash features.
Although hash features converge quickly and can capture fine-grained details, we observe that directly using hash features at all resolutions leads to noise and floating artifacts with sparse input views.
This is %due to 
\yh{because} the expressive ability of multi-resolution hash features is too powerful. They can easily overfit to the few-shot training views, causing the optimized SDF to fall into local minima.

Inspired by FreeNeRF~\cite{freenerf}, we assume that the low-resolution hash features encode a coarse geometric shape, while the high-resolution features represent high-frequency information.
Hence, to avoid quickly overfitting and unwanted artifacts, we adopt a carefully designed training scheme that progressively activates hash features.
In the early stages of training, we only use low-resolution hash features to generate overly smooth geometries. In the later stages, we gradually use higher-resolution hash features to compensate for fine-grained details.
This scheme can be achieved by applying a progressive activation mask:
\begin{equation}
\begin{split}
\label{eqn:progressive}
\widetilde{\gamma}(\bm{x}, \psi) & = m(\psi) \odot \gamma(\bm{x}), \\
m(\psi) = (m_0( & \psi), m_1(\psi), ..., m_L(\psi)),
\end{split}
\end{equation}
where $m_i(\psi) = \bm{I}[i \leq \psi]$, and $\psi$ controls the bandwidth of the hash encoding. In practice, we set $\psi = \frac{i}{\Theta}L$, where $i$ is the current training iteration, $\Theta$ is a predefined threshold, and $L$ is the resolution level of hash features.
Our progressive activation scheme shares some similarities with the coarse-to-fine strategy used in~\citet{neuralangelo}. Different from theirs, our scheme focuses on solving the severe overfitting problem under sparse-view SDF optimization.
This scheme plays a vital role in our framework, as we aim to achieve high-quality reconstruction without artifacts, and a well-optimized reconstruction module also provides more informative guidance to the view planning module.

\revise{
\subsection{Directional Hessian Loss}
To produce reasonable geometry, recent works apply an Eikonal loss $\mathcal{L}_{eik}$~\cite{eikonal} for regularizing the gradient of SDF to be close to one.
However, we find that $\mathcal{L}_{eik}$ only considers the first-order gradient of SDF and is difficult to provide sufficient regularization, especially under challenging sparse-view settings.
A higher-order gradient can provide a stronger constraint.
Some studies~\cite{reg-surf, neuralangelo} compute the second-order Hessian matrix and directly regularize the matrix norm to zero.
Yet, we notice that the second-order gradient of SDF is not necessarily zero everywhere in space. Instead, the directional derivative of the SDF gradient along its normal direction should be zero due to the parallelism of adjacent SDF level sets near the surface.
Therefore, we propose a directional Hessian loss $\mathcal{L}_{dir}$, which can constrain the second-order gradient more precisely:
\begin{equation}
\label{eqn:dir_loss}
\mathcal{L}_{dir} = exp(-\delta \cdot |f_g(\bm{x})|) \cdot \frac{|\nabla f_g(\bm{x})| - |\nabla f_g(\bm{x} + \epsilon \cdot \frac{\nabla f_g(\bm{x})}{\left \|\nabla f_g(\bm{x}) \right \|})|}{\epsilon},
\end{equation}
where $\nabla$ is the gradient operator, and we set the step size $\epsilon$ to equal to the minimum grid size of hash encoding; $exp(-\delta \cdot |f_g(\bm{x})|)$ is a radial basis function (RBF) with a hyperparameter $\delta$, which encourages the loss to focus more on regions near the surface.
Our proposed loss $\mathcal{L}_{dir}$ can serve as a smoothness constraint by regularizing the inconsistency of the SDF gradient, thus facilitating the ill-posed SDF optimization under sparse views.
}

\begin{small}
\begin{table*}[tb]
\centering
\setlength{\abovecaptionskip}{3pt}  
\setlength{\belowcaptionskip}{0pt}
\caption{Quantitative results on Blender dataset (bold means best, underline means second best). We report the PSNR and SSIM.}
\renewcommand\arraystretch{0.8}
\setlength{\tabcolsep}{4.85pt}
\begin{tabular}{c|c|cc|cc|cc|cc|cc|cc}
\toprule
\multirow{2.5}{*}{\textbf{Methods}}& \multirow{2.5}{*}{\textbf{Time}} & \multicolumn{2}{c|}{\textbf{Mic}} & \multicolumn{2}{c|}{\textbf{Hotdog}} & \multicolumn{2}{c|}{\textbf{Chair}} & \multicolumn{2}{c|}{\textbf{Ficus}} & \multicolumn{2}{c|}{\textbf{Materials}} & \multicolumn{2}{c}{\textbf{mean}} \\
\cline{3-14}
\rule{0pt}{8pt}
& & PSNR$\uparrow$ & SSIM$\uparrow$ & PSNR$\uparrow$ & SSIM$\uparrow$ & PSNR$\uparrow$ & SSIM$\uparrow$ & PSNR$\uparrow$ & SSIM$\uparrow$ & PSNR$\uparrow$ & SSIM$\uparrow$ & PSNR$\uparrow$ & SSIM$\uparrow$ \\
\midrule
NeuS (\textit{random}) & \multirow{3}{*}{91min} & 18.65 & 0.871 & 26.55 & 0.928 & 23.64 & 0.886 & 17.23 & 0.830 & 18.45 & 0.789 & 20.90 & 0.861 \\
NeuS (\textit{cluster}) & & 19.83 & 0.879 & 27.26 & 0.933 & 25.30 & 0.901 & 17.58 & 0.831 & 18.74 & 0.794 & 21.74 & 0.868 \\
NeuS (\textit{farthest}) & & 19.38 & 0.873 & 21.93 & 0.896 & 25.32 & 0.903 & 17.71 & 0.834 & 17.92 & 0.780 & 20.45 & 0.857 \\
\midrule
Neuralangelo (\textit{random}) & \multirow{3}{*}{518min} & 26.85 & \uline{0.952} & 23.04 & 0.919 & \uline{26.85} & \uline{0.933} & 17.38 & 0.835 & 18.60 & 0.801 & 22.54 & 0.888 \\
Neuralangelo (\textit{cluster}) & & 26.60 & 0.947 & 25.08 & 0.928 & \textbf{26.96} & \textbf{0.934} & 17.57 & 0.833 & \textbf{20.41} & 0.819 & 23.32 & 0.892 \\
Neuralangelo (\textit{farthest}) & & 26.64 & 0.950 & 23.69 & 0.905 & 26.22 & 0.928 & 17.65 & 0.831 & 16.12 & 0.775 & 22.06 & 0.878 \\
\midrule
MonoSDF (\textit{random}) & \multirow{3}{*}{386min} & 23.58 & 0.928 & 24.99 & 0.919 & 20.69 & 0.848 & 17.59 & 0.852 & 19.40 & 0.827 & 21.25 & 0.875 \\
MonoSDF (\textit{cluster}) & & 23.29 & 0.927 & 25.24 & 0.924 & 23.61 & 0.882 & 19.52 & 0.863 & 19.65 & \textbf{0.838} & 22.26 & 0.887 \\
MonoSDF (\textit{farthest}) & & 22.99 & 0.925 & 23.39 & 0.902 & 23.38 & 0.883 & 18.89 & 0.861 & 19.67 & 0.830 & 21.66 & 0.880 \\
\midrule
Ours (\textit{random}) & \multirow{4}{*}{10min} & \uline{27.65} & 0.951 & 26.97 & \uline{0.935} & 25.16 & 0.910 & \uline{22.36} & \textbf{0.903} & 18.12 & 0.804 & 24.05 & 0.901 \\
Ours (\textit{cluster}) & & 27.43 & \uline{0.952} & \uline{27.85} & 0.929 & 26.33 & 0.919 & 22.13 & 0.898 & 19.43 & 0.818 & \uline{24.63} & \uline{0.903} \\
Ours (\textit{farthest}) & & 26.45 & 0.950 & 26.24 & 0.917 & 25.46 & 0.911 & 22.30 & 0.895 & 18.25 & 0.807 & 23.74 & 0.896 \\
Ours (\textit{planning}) & & \textbf{27.87} & \textbf{0.956} & \textbf{28.51} & \textbf{0.936} & 26.54 & 0.925 & \textbf{22.59} & \uline{0.899} & \uline{20.23} & \uline{0.832} & \textbf{25.15} & \textbf{0.910} \\
\bottomrule
\end{tabular}
\label{tab:blender_comp}
\end{table*}
\end{small}

\subsection{Overall Loss}
 \label{sec:loss}
The total loss $\mathcal{L}$ used to optimize our model is:
\begin{equation}
\label{eqn:overall_loss}
\mathcal{L} = \mathcal{L}_{rgb} + w_{norm} \mathcal{L}_{norm} + w_{eik} \mathcal{L}_{eik} + w_{dir} \mathcal{L}_{dir},
\end{equation}
where the RGB loss $\mathcal{L}_{rgb} = \frac{1}{| \mathcal{R} |} \sum_{\bm{r} \in \mathcal{R}} {\left \| C(\bm{r}) - \hat{C}(\bm{r}) \right \|_1}$ minimizes the difference between the rendered pixel $C(\bm{r})$ and ground-truth pixel $\hat{C}(\bm{r})$ for sampled ray $\bm{r}$. Here, $\mathcal{R}$ denotes the set of rays in each batch.
To further regularize the surface, we also predict the surface normal using Omnidata~\cite{omnidata} and apply a normal loss.
The normal loss $\mathcal{L}_{norm} = \frac{1}{| \mathcal{R} |} \sum_{\bm{r} \in \mathcal{R}} {\left \| 1 -  N(\bm{r})^\top \hat{N}(\bm{r}) \right \|_1}$ constrains the rendered normal vector $N(\bm{r})$ to be consistent with the predicted pseudo ground-truth normal vector $\hat{N}(\bm{r})$.

\section{Experiments}
\noindent \textbf{\textit{Implementation details.}} We adopt a 12-level hash encoding, with each hash entry having a channel size of 2. The total number of hash entries is $2^{19}$. 
We progressively activate hash features with finer resolutions until $\Theta=10,000$ iterations.
We use 1 hidden layer with 64 hidden units for the SDF decoder and 2 hidden layers with 64 hidden units for the color MLP. 
We first train our framework using 3 clustered views for 1,000 iterations as initialization. Then, we add a new view using the proposed strategy every 1,000 iterations until the number of training views reaches a target value.
The loss weights are set to $w_{norm} = w_{dir} = 0.05$, and $w_{eik} = 0.1$.

\ 

\noindent \textbf{\textit{Datasets.}} 
We simulate our problem setting and conduct experiments on three datasets. Candidate viewpoints are limited to viewpoints of images in these datasets.
\yh{Similar to some previous works~\cite{neus, sparseneus}, we use representative scenes covering different aspects (\textit{i.e.}, materials, appearances, and geometries) for evaluation.
}
Specifically, we use 10 scenes from the DTU dataset~\cite{dtu} and 5 challenging scenes \revise{(\textit{i.e.}, Bear, Cattle, Clock, Man, Sculpture)} from the BlendedMVS dataset~\cite{bmvs} to evaluate the surface reconstruction quality.
We further test on 5 scenes \revise{(we exclude scenes containing semi-transparent, hollow objects or fluids, which are not suitable for SDF-based surface reconstruction methods)} from the Blender dataset~\cite{nerf} to evaluate the novel view synthesis quality.
For each dataset, we use only 10\% of dense views for reconstruction.
Concretely, we use 6 and 8 views for DTU and Blender, and 8-12 views for BlendedMVS.

\ 

\noindent \textbf{\textit{Baselines.}} \yh{Our proposed view planning and reconstruction modules can be flexibly switched out in a modular manner against other exiting methods. For the reconstruction module, we} %We 
compare %our \textit{PVP-Recon} 
with the following baselines: (1) The state-of-the-art surface reconstruction methods, including NeuS~\cite{neus} and Neuralangelo~\cite{neuralangelo}; (2) The state-of-the-art sparse-view methods, including generalization-based SparseNeuS~\cite{sparseneus}, VolRecon~\cite{volrecon} and regularization-based MonoSDF~\cite{monosdf}.
Because these methods use fixed input views before training, we apply three policies to select input views for them: \textit{random sampling}, \textit{cluster sampling}, \textit{farthest sampling}. 
Exceptionally, for generalization-based methods, we use predefined views from their original methods for comparison, as their performance drops significantly when using other views as input.
\yh{For the view planning module, we replace our proposed view planning strategy with different strategies introduced by \citet{active-3d}, NeurAR \cite{neurar}, and NeU-NBV \cite{neu-nbv} to show its effectiveness.}
%In ablation study, we also replace our proposed view planning strategy with different strategies introduced by~\cite{active-3d} and~\cite{neurar} to compare the effectiveness. 

\begin{figure*}[t]% specify a combination of t, b, p, or h for top, bottom, on its own page, or here
  \centering % avoid the use of \begin{center}...\end{center} and use \centering instead (more compact)
  \setlength{\abovecaptionskip}{4pt}  
  \setlength{\belowcaptionskip}{0pt}
  \includegraphics[width=0.93\linewidth]{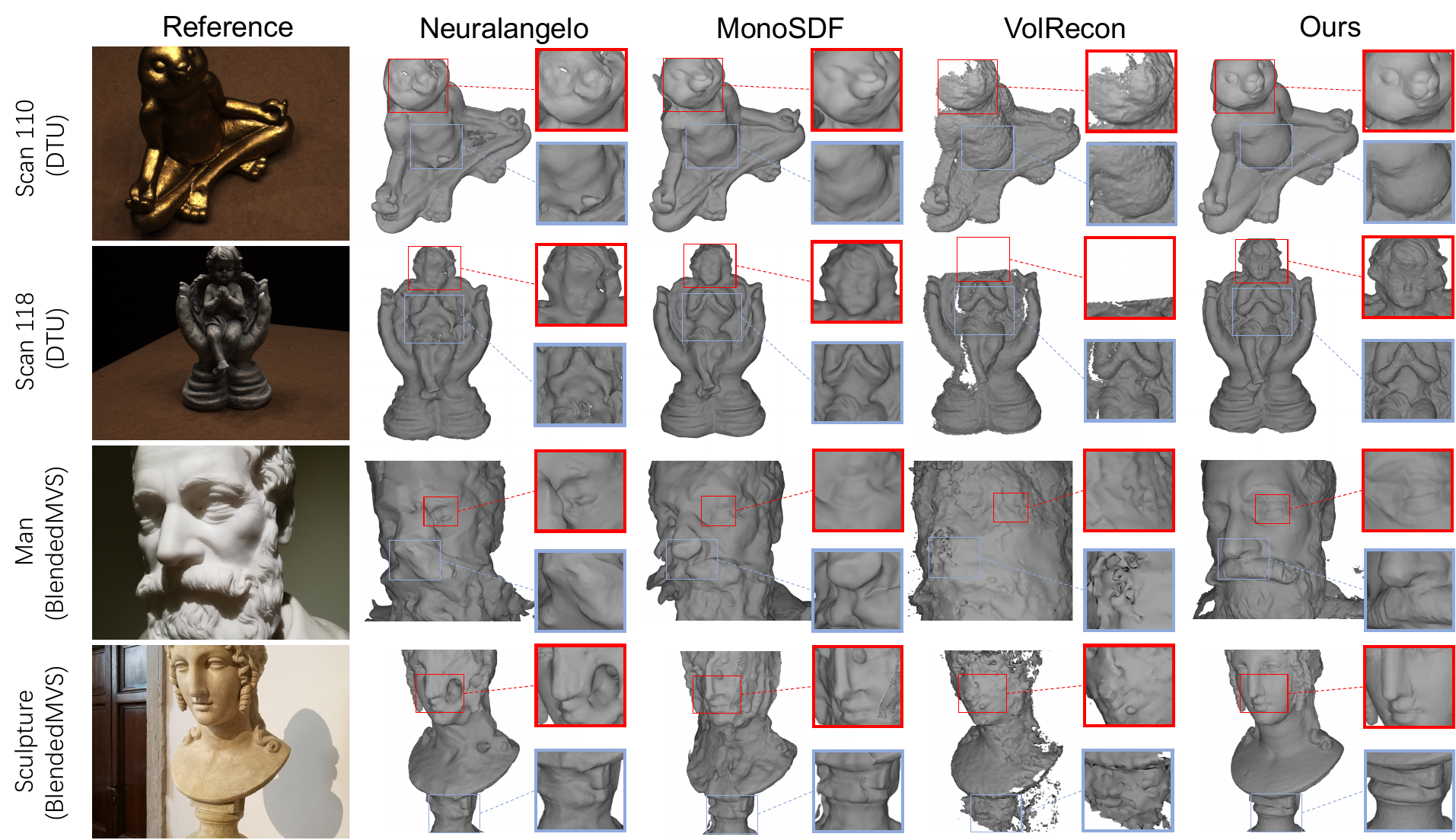}
  \caption{%
    Comparison of surface reconstruction on DTU and BlendedMVS datasets. Our method generates the most accurate and detailed results.
  }
  \label{fig:vis_comp_mesh}
  \Description{vis_comp_mesh}
\end{figure*}

\begin{figure*}[t]% specify a combination of t, b, p, or h for top, bottom, on its own page, or here
  \centering % avoid the use of \begin{center}...\end{center} and use \centering instead (more compact)
  \setlength{\abovecaptionskip}{4pt}  
  \setlength{\belowcaptionskip}{0pt}
  \includegraphics[width=0.93\linewidth]{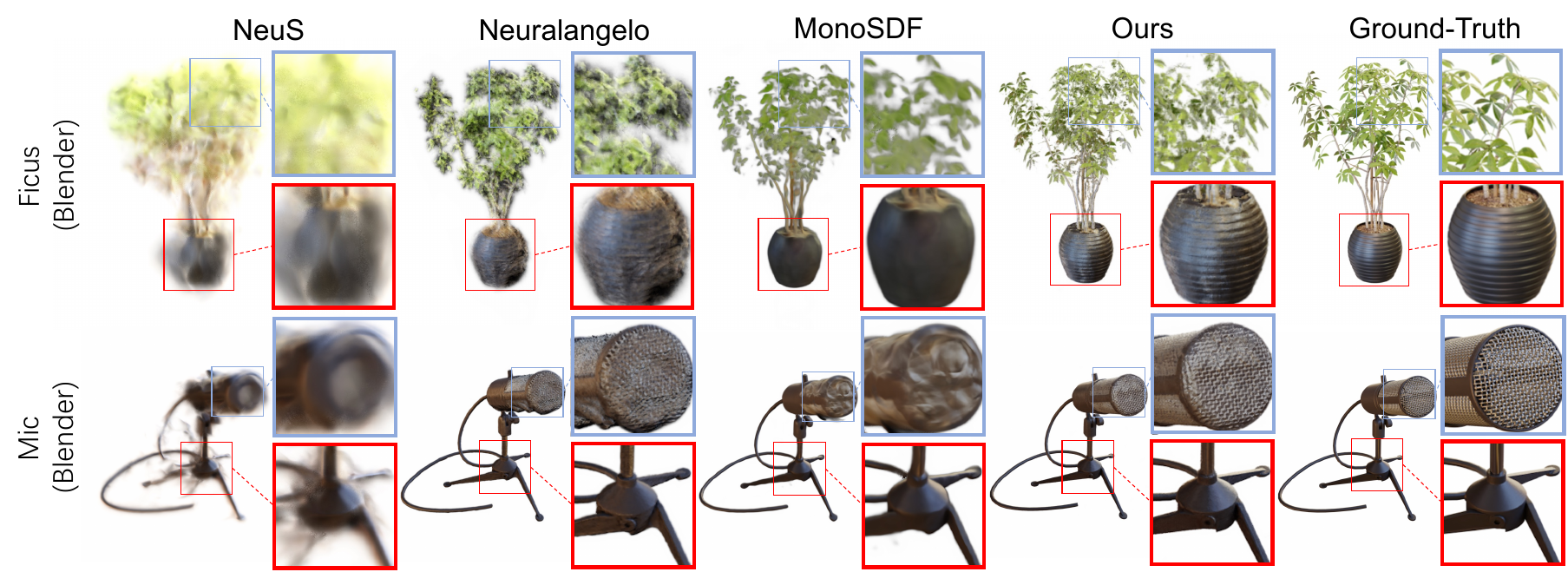}
  \caption{%
    Comparison of image rendering on Blender dataset. Our method can generate high-quality renderings with richer details.
  }
  \label{fig:vis_comp_render}
  \Description{vis_comp_render}
  \vspace{-0.1cm}
\end{figure*}

\subsection{Quantitative Evaluation} \label{sec:quanexp}
\ys{To evaluate our reconstruction module, we use the same three policies to select input views and compare with baselines.}
In Table \ref{tab:dtu_comp}, we report Chamfer distance on the DTU dataset to measure the reconstruction accuracy. 
The object masks are used to remove the background for proper evaluation~\cite{sparseneus}.
To further evaluate image synthesis quality, we report the peak signal-to-noise ratio (PSNR) and structural similarity index measure (SSIM) on the Blender dataset (see Table \ref{tab:blender_comp}).
On average, \textit{PVP-Recon} achieves the lowest Chamfer distance and the highest PSNR and SSIM with significantly less training time.
\ys{The results indicate that our reconstruction module generally outperforms existing works in terms of surface reconstruction and image rendering.
We also notice that our reconstruction quality can be further improved after applying the view planning module (\textit{i.e.}, \textit{planning}).
}

For further evaluation of the view planning module, we compare our proposed strategy with three representative strategies in the field of robotics.
\citet{active-3d} calculate the density entropy along each ray and use this entropy as a measure for view planning.
Yet, density struggles to provide the most valuable information for geometric reconstruction.
\ys{Also, calculating entropy for each ray results in a heavy computational burden.
NeurAR \cite{neurar} and NeU-NBV \cite{neu-nbv} model the emitted radiance as Gaussian distributions and plans subsequent views using the distribution variance.}
Nevertheless, expanding the radiance from a specific value to a distribution induces randomness, which makes the optimization difficult and degrades the reconstruction quality.
In contrast, our strategy is simple yet effective as it directly verifies the multi-view consistency via image warping.
\ys{We implement and incorporate these strategies into our framework and conduct experiments on the DTU dataset. The entropy, variance maps, and warping scores are all calculated at 150 $\times$ 200 pixel resolution.}
\ys{Table \ref{tab:strategy} shows the averaged Chamfer distance and \secondary{averaged} running time for each round of view planning.}
\ys{Our strategy consistently outperforms other strategies under different input views with comparable running time.
}

\begin{figure*}[t]% specify a combination of t, b, p, or h for top, bottom, on its own page, or here
  \centering % avoid the use of \begin{center}...\end{center} and use \centering instead (more compact)
  \setlength{\abovecaptionskip}{6pt}  
  \setlength{\belowcaptionskip}{0pt}
  \includegraphics[width=0.88\linewidth]{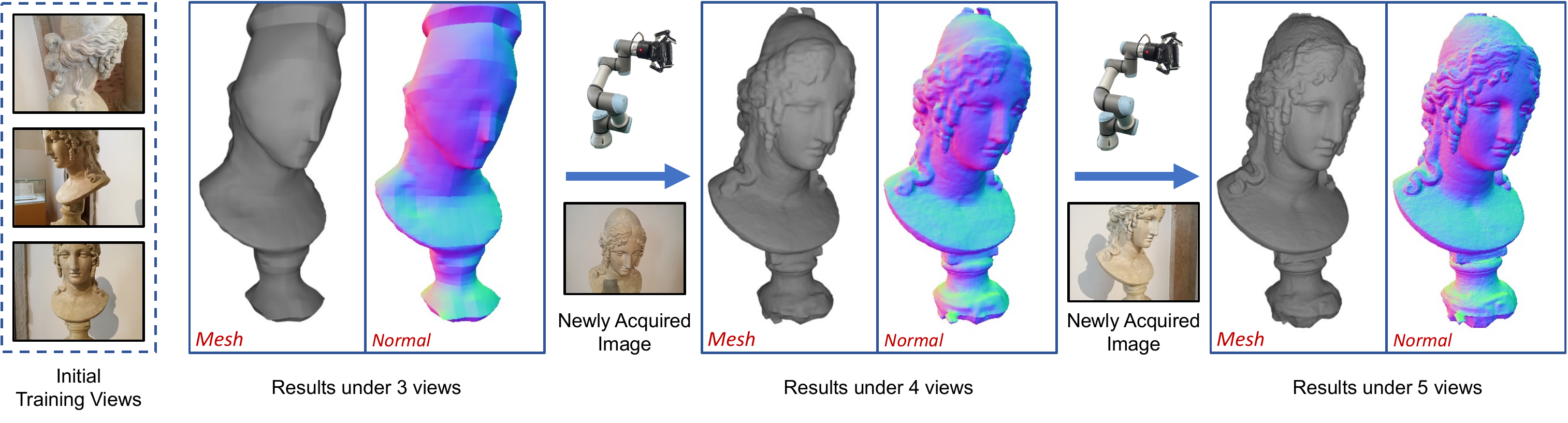}
  \caption{%
    We show how the reconstructed mesh \revise{of BlendedMVS Sculpture} changes as new input views are progressively added. Note that the newly added views provide beneficial information that makes the reconstruction result more precise and detailed.
  }
  \label{fig:progress}
  \Description{progress}
  \vspace{-0.1cm}
\end{figure*}

\begin{figure*}[t]% specify a combination of t, b, p, or h for top, bottom, on its own page, or here
  \centering % avoid the use of \begin{center}...\end{center} and use \centering instead (more compact)
  \setlength{\abovecaptionskip}{5pt}
  \setlength{\belowcaptionskip}{1pt}
  \includegraphics[width=0.9\linewidth]{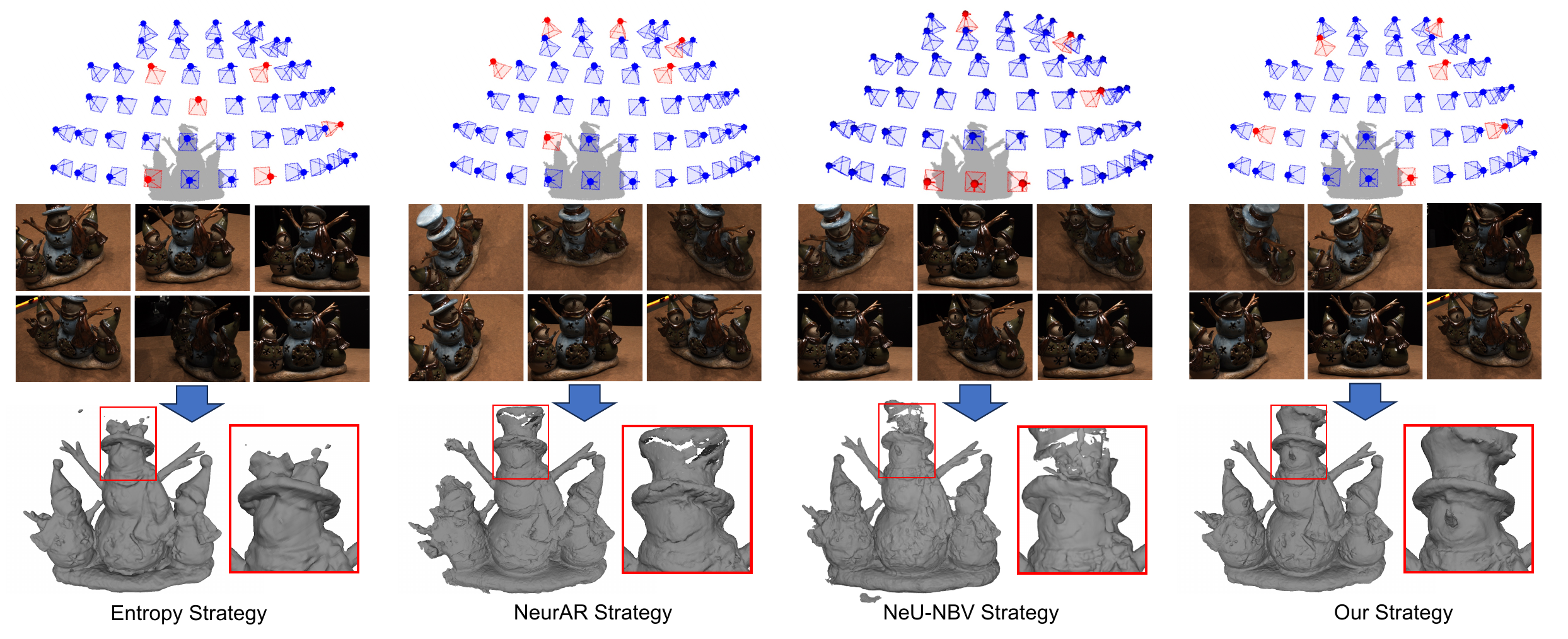}
  \caption{%
    We visualize the selected cameras, corresponding RGB inputs, and final recovered meshes \revise{(DTU scan69)} of different view planning strategies.
  }
  \label{fig:strategy_comp}
  \Description{strategy_comp}
  \vspace{-0.2cm}
\end{figure*}

\begin{table}[t]
\centering
\setlength{\abovecaptionskip}{4pt}
\setlength{\belowcaptionskip}{0pt}
\caption{\revise{Averaged Chamfer distance and running time of different view planning strategies on 10 scenes of the DTU dataset.}}
\setlength{\tabcolsep}{4.0pt}
\begin{tabular}{c|cccc}
\toprule
\textbf{Strategy Type} & 4 views & 5 views & 6 views & Time \\
\midrule
Entropy~\cite{active-3d} & 1.20 & 1.07 & 0.93 & 59s \\
NeurAR~\cite{neurar} & 1.51 & 1.36 & 1.23 & \textbf{6.5s} \\
\revise{NeU-NBV~\cite{neu-nbv}} & \revise{1.35} & \revise{1.25} & \revise{1.12} & \revise{11.1s} \\
Ours & \textbf{0.99} & \textbf{0.92} & \textbf{0.81} & 7.8s \\
\bottomrule
\end{tabular}
\label{tab:strategy}
\end{table}

\subsection{Qualitative Evaluation}

We visualize the reconstructed meshes and conduct qualitative comparisons (shown in Figure \ref{fig:vis_comp_mesh}).
For each baseline, we choose the best results from three view selection policies.
Neuralangelo~\cite{neuralangelo} and MonoSDF~\cite{monosdf} struggle to generate accurate geometries for textureless regions (Scan 110) or delicate structures (Scan 118).
VolRecon~\cite{volrecon} generates missing or noisy results.
Moreover, baselines fail to generate good results on challenging scenes of the BlendedMVS dataset.
In contrast, \textit{PVP-Recon} can reconstruct both smooth surfaces and detailed structures.

Although our primary goal is surface reconstruction, we also evaluate rendering quality using the Blender dataset, which provides a separate test set for evaluating novel view synthesis.
Figure \ref{fig:vis_comp_render} shows that \textit{PVP-Recon} also outperforms baselines in terms of rendering quality.
NeuS \cite{neus} and MonoSDF \cite{monosdf} synthesize blurry images because of the low-frequency bias of the MLP networks.
Neuralangelo \cite{neuralangelo} captures more details but produces incorrect textures in complex areas. This is due to its ill-posed optimization of hash features under sparse views.
By utilizing our proposed training scheme and the directional Hessian loss $\mathcal{L}_{dir}$, our model is better optimized and recovers sharper and more visually appealing rendering results.

\ys{
In Figure \ref{fig:progress}, we show the evolution process of the reconstructed mesh under progressively added input views. 
The recovered mesh becomes more precise and richer in details, as our view planning module can supplement the most informative views for surface reconstruction.
We also visualize the selected cameras, corresponding RGB images, and final reconstruction results of different planning strategies for comparison in Figure \ref{fig:strategy_comp}.
Input views planned by entropy-based strategy \cite{active-3d} tend to neglect edge regions, leading to incomplete surfaces of the target object.
Variance-based strategies (\textit{i.e.}, NeurAR \cite{neurar} and NeU-NBV \cite{neu-nbv}) can plan reasonable input views, but introduce randomness into the surface optimization process, making the recovered meshes bumpy.
Our warping-based strategy achieves better capture coverage and results in high-quality and delicate surfaces.
}

\begin{table}[t]
\centering
\setlength{\abovecaptionskip}{3pt}
\setlength{\belowcaptionskip}{0pt}
\caption{\revise{Ablation results (averaged Chamfer distance) on the DTU dataset.}}
\setlength{\tabcolsep}{9.5pt}
\begin{tabular}{c|ccc}
\toprule
\textbf{Configs} & \textit{w/o prog} & \textit{w/o $\mathcal{L}_{dir}$} & \textit{full setup} \\
\midrule
farthest sampling & 1.53 & 1.16 & \textbf{1.01} \\
view planning & 1.45 & 0.91 & \textbf{0.81} \\
\bottomrule
\end{tabular}
\label{tab:ablation}
\end{table}

\subsection{Ablation Study} 
\begin{figure}[t]% specify a combination of t, b, p, or h for top, bottom, on its own page, or here
  \centering % avoid the use of \begin{center}...\end{center} and use \centering instead (more compact)
  \setlength{\abovecaptionskip}{4pt}  
  \setlength{\belowcaptionskip}{0pt}
  \includegraphics[width=0.9\columnwidth]{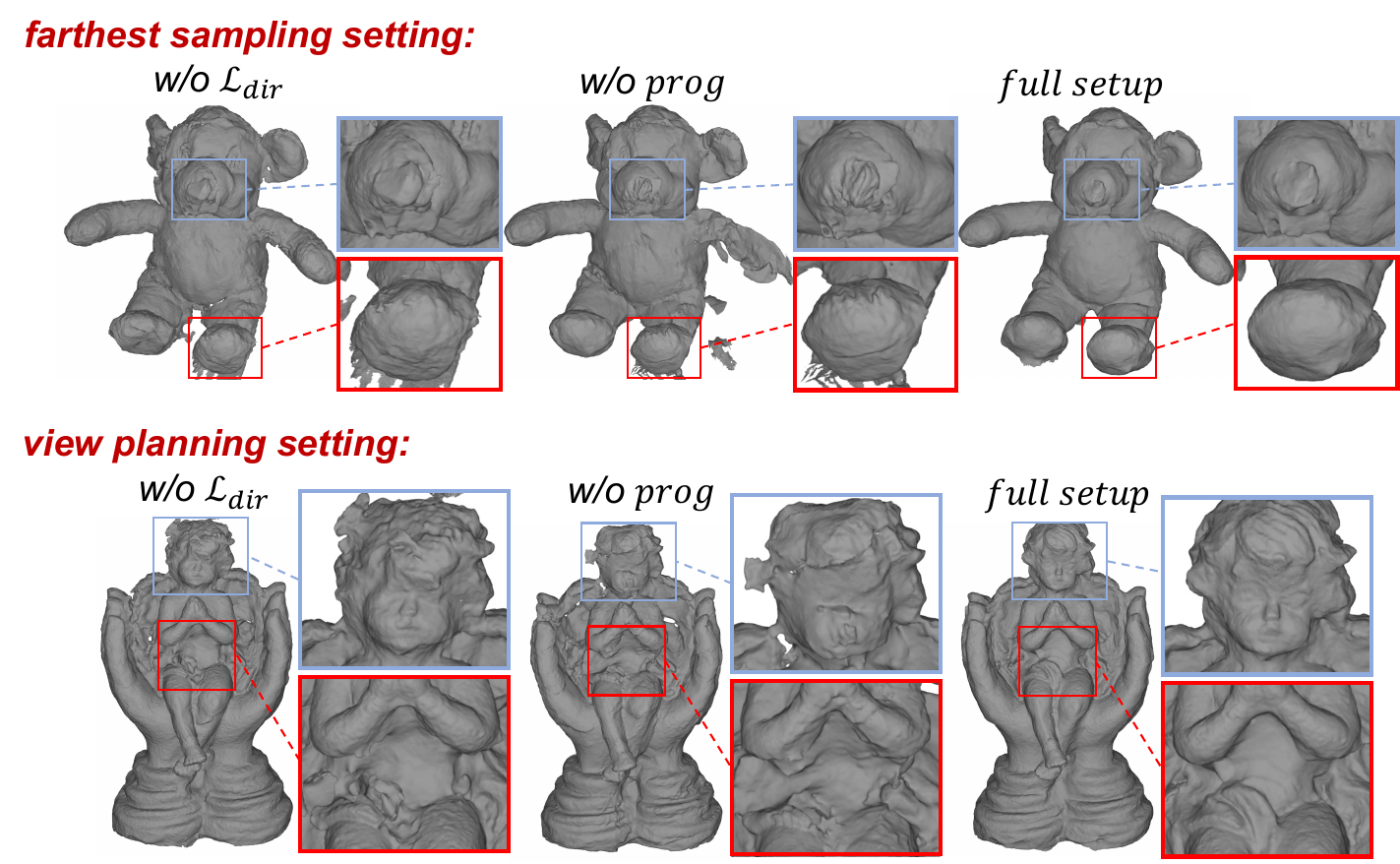}
  \caption{%
    Ablation visualizations \revise{(DTU scan105 and scan118)}. Note that Artifacts will appear when any of our proposed components are removed.
  }
  \label{fig:ablation_vis}
  \Description{ablation_vis}
  \vspace{-0.4cm}
\end{figure}

We conduct more ablation experiments on the DTU dataset.
We remove the following key components separately: (1) progressive training scheme (\textit{w/o prog}); (2) directional Hessian loss (\textit{w/o} $\mathcal{L}_{dir}$). 
\ys{Note that we evaluate the effectiveness of these components both with and without (\textit{i.e.}, \textit{farthest sampling}) the view planning module, to avoid entangling the evaluation with changes in the input views.} 
Table \ref{tab:ablation} and Figure \ref{fig:ablation_vis} show the ablation results.
\ys{In both settings, the full setup achieves the best results, and the reconstruction quality deteriorates when either component is removed.}
Specifically, the progressive scheme prevents the reconstruction optimization from quickly falling into local minima. Without this scheme (\textit{w/o prog}), the model fails to capture scene details and produces messy geometries.
Intuitively, our proposed loss $\mathcal{L}_{dir}$ constrains the gradient of SDF and acts as a consistency prior by regularizing abrupt changes in surface curvature.
Therefore, the reconstructed meshes are non-smooth and inconsistent when this loss is removed (\textit{w/o} $\mathcal{L}_{dir}$).
As shown in Figure \ref{fig:ablation_vis}, our full setup generates the most accurate and complete surfaces with fine-grained details.

\begin{figure*}[t]% specify a combination of t, b, p, or h for top, bottom, on its own page, or here
  \centering % avoid the use of \begin{center}...\end{center} and use \centering instead (more compact)
  \setlength{\abovecaptionskip}{4pt}  
  \setlength{\belowcaptionskip}{0pt}
  \includegraphics[width=0.93\linewidth]{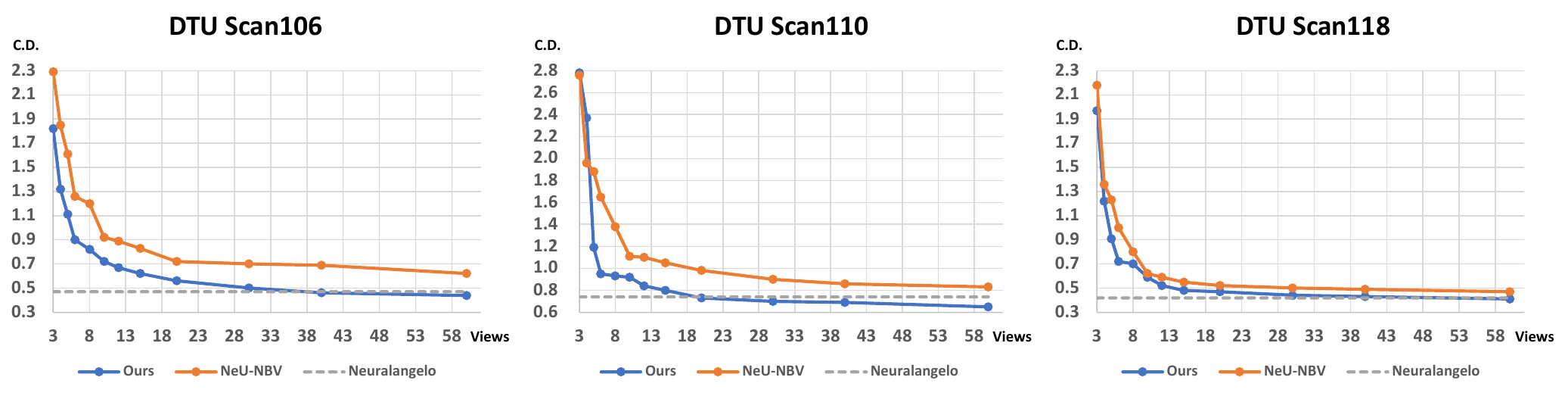}
  \caption{
    \revisenew{Reconstruction results (the Chamfer distance, C.D. $\downarrow$ ) of ours and \cite{neu-nbv,neuralangelo} using different numbers of input views on DTU dataset. } %We report the Chamfer distance (C.D.) $\downarrow$ variation curves.}
  }
  \label{fig:view_number}
  \Description{view_number}
  \vspace{-0.1cm}
\end{figure*}

\begin{figure}[t]% specify a combination of t, b, p, or h for top, bottom, on its own page, or here
  \centering % avoid the use of \begin{center}...\end{center} and use \centering instead (more compact)
  \setlength{\abovecaptionskip}{6pt}  
  \setlength{\belowcaptionskip}{0pt}
  \includegraphics[width=0.9\columnwidth]{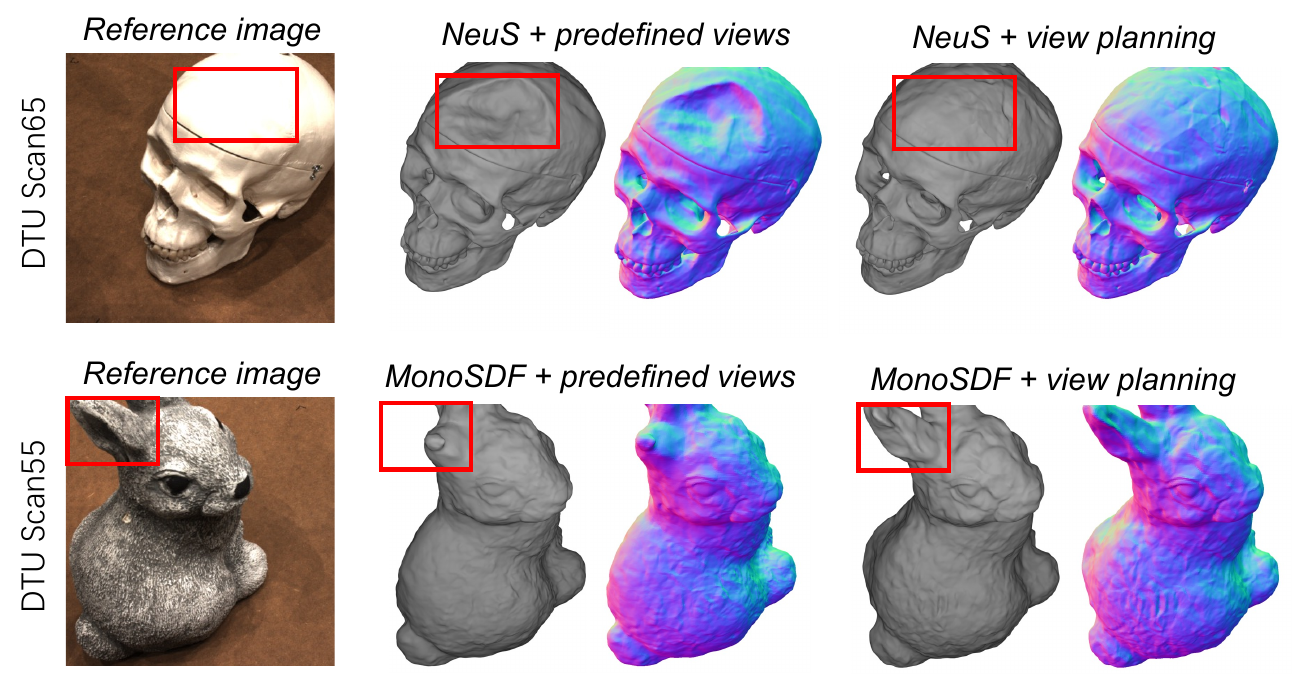}
  \caption{%
    Our proposed view planning module can be flexibly incorporated into different 3D reconstruction methods and help to reconstruct more accurate and complete surfaces.
  }
  \label{fig:flexibility}
  \Description{flexibility}
  \vspace{-0.2cm}
\end{figure}

\revisenew{
\subsection{Different Numbers of Input Views}
We evaluate the reconstruction quality under different numbers of input views.
Figure \ref{fig:view_number} shows the Chamfer distance (C.D.) variation curves for three representative scenes in the DTU dataset. With fewer input views, the Chamfer distance decreases significantly each time a new image is added. 
Our \textit{PVP-Recon} generally outperforms NeU-NBV \cite{neu-nbv} strategy under different numbers of views.
We also report the reconstruction results of Neuralangelo \cite{neuralangelo} using all dense images (64 views) of a scene. Note that we can achieve comparable or even better results with fewer input images, which demonstrates the effectiveness of our system.
}

\subsection{Flexibility of the View Planning Module}
\ys{
Our warping-based view planning module is flexible and can be combined with %other 
different 3D reconstruction methods. 
We incorporate the view planning module into NeuS \cite{neus} and MonoSDF \cite{monosdf} framework.
Figure \ref{fig:flexibility} and Table \ref{tab:flexibility} show that, compared with predefining a fixed set of input views using the \textit{cluster sampling}, our incorporated view planning module can further help NeuS and MonoSDF to reconstruct more accurate and complete surfaces, and achieve lower Chamfer distance.
}

\begin{table}[t]
\centering
\setlength{\abovecaptionskip}{4pt}
\setlength{\belowcaptionskip}{0pt}
\caption{%Combining other 3D reconstruction methods with 
Integrating our view planning module into different 3D reconstruction methods can achieve \revise{lower averaged Chamfer distance on the DTU dataset} than predefining a fixed set of views before training.}
\setlength{\tabcolsep}{5pt}
\begin{tabular}{c|cc}
\toprule
\textbf{Methods} & predefined views & view planning \\
\midrule
NeuS \cite{neus} & 1.27 & \textbf{1.15} \\
MonoSDF \cite{monosdf} & 1.44 & \textbf{1.36} \\
\bottomrule
\end{tabular}
\label{tab:flexibility}
\vspace{-0.2cm}
\end{table}

\begin{figure}[t]% specify a combination of t, b, p, or h for top, bottom, on its own page, or here
  \centering % avoid the use of \begin{center}...\end{center} and use \centering instead (more compact)
  \setlength{\abovecaptionskip}{4pt}  
  \setlength{\belowcaptionskip}{0pt}
  \includegraphics[width=0.9\columnwidth]{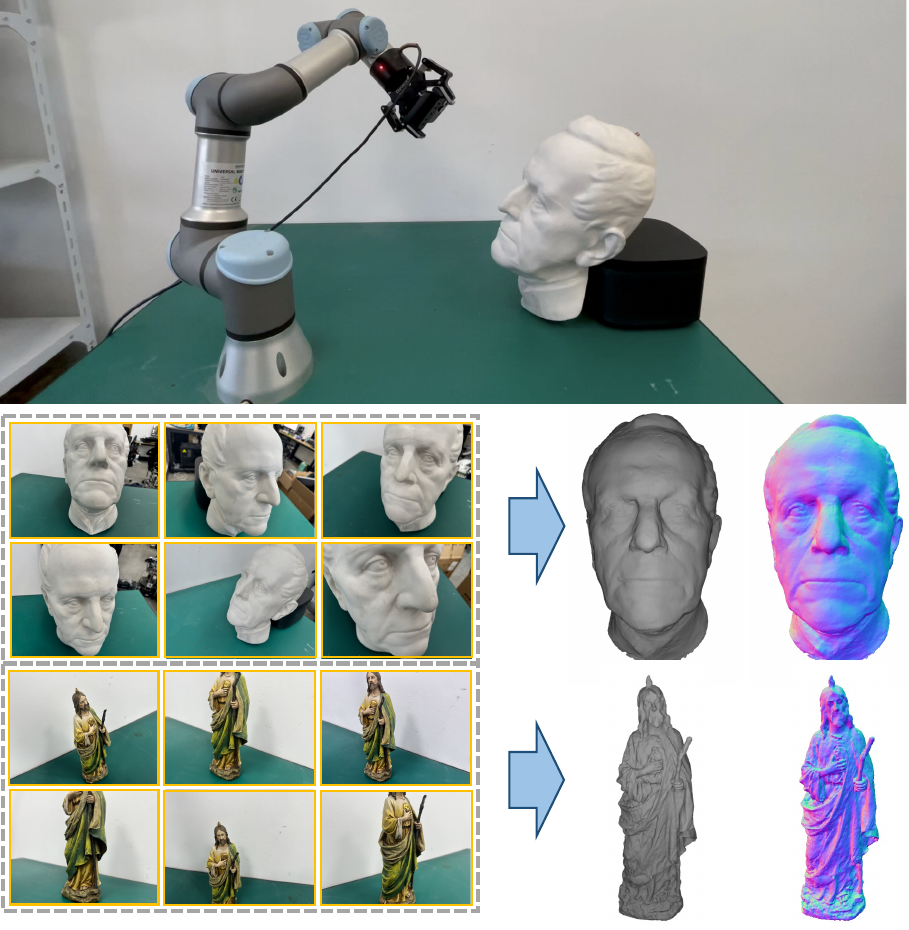}
  \caption{%
    The meshes reconstructed by \textit{PVP-Recon} from sparse-view images progressively captured by a robotic arm.
  }
  \label{fig:robot}
  \Description{robot}
  \vspace{-0.2cm}
\end{figure}

\subsection{Real-World Robotic Application}
\textit{PVP-Recon} alternately optimizes the surface and plans subsequent input views based on current optimization status.
Therefore, \textit{PVP-Recon} is well suited for active reconstruction in the field of robotics and can be applied in practice.
In Figure \ref{fig:robot}, we show meshes reconstructed by \textit{PVP-Recon} from sparse images progressively captured by a robotic arm. \textit{PVP-Recon} can still generate accurate and high-quality 3D mesh surfaces in real-world robotic scenarios.

\section{Limitations and Conclusion}
\textbf{Limitations.} Currently, \textit{PVP-Recon} takes 8 seconds each time for view planning and 10 minutes for the overall reconstruction. Future direction includes achieving further acceleration with CUDA implementation for applications with high requirements on real-time performance.
\ys{Moreover, \textit{PVP-Recon} now focuses on object-level scene reconstruction. In the future, we aim to combine our method with mobile robots or drones to achieve full 3D reconstruction of large, unbounded, and non-object-centric scenarios.}

~

\noindent \textbf{Conclusion.} In this paper, we propose \textit{PVP-Recon}, a novel sparse-view surface reconstruction system that progressively plans the next best views to form an optimal set of input images.
Specifically, we design a scoring strategy that checks the multi-view consistency to seek the most informative images for further training.
To stabilize the implicit surface optimization under sparse views, we also introduce a progressive training scheme and a directional Hessian loss.
Extensive experiments on three datasets show that our method outperforms previous reconstruction baselines and active planning strategies.
Furthermore, we demonstrate that \textit{PVP-Recon} is well suited for active reconstruction task in the field of robotics, and our proposed view planning module can also be incorporated into other frameworks to facilitate better surface reconstruction.

\section*{Acknowledgments}
This work was supported by Beijing Science and Technology plan project (Z231100005923029), the Natural Science Foundation of China (62332019, U2336214), and The Talent Fund of Beijing Jiaotong University (2023XKRC045).

\balance

\normalem
% Bibliography
\bibliographystyle{ACM-Reference-Format}
\bibliography{sample-bibliography}

\begin{figure*}[t]% specify a combination of t, b, p, or h for top, bottom, on its own page, or here
  \centering % avoid the use of \begin{center}...\end{center} and use \centering instead (more compact)
  \setlength{\abovecaptionskip}{2pt}  
  \setlength{\belowcaptionskip}{0pt}
  \includegraphics[width=0.85\linewidth]{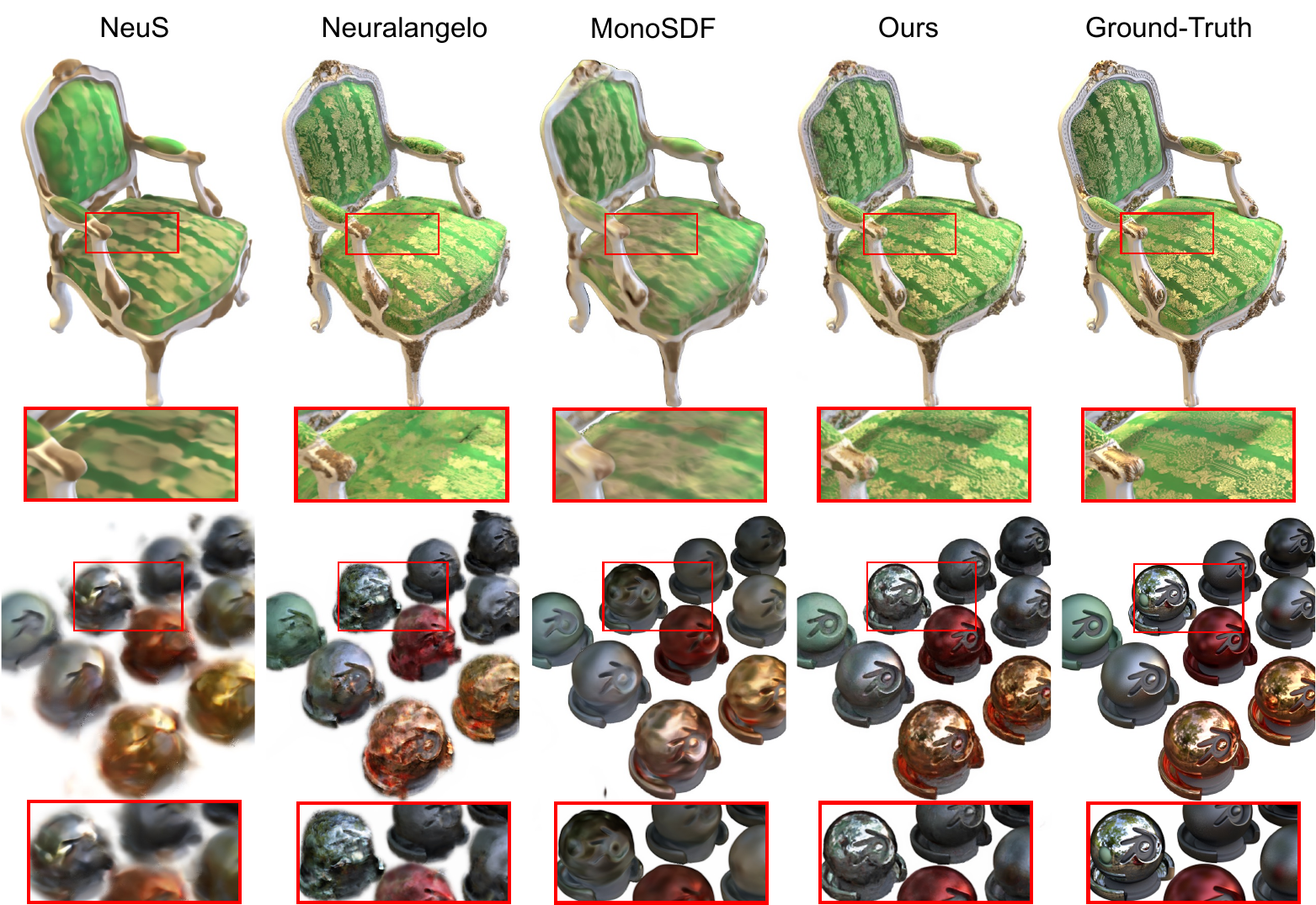}
  \caption{%
    Additional comparison results of image rendering on Blender. Our approach can generate high-quality renderings with richer details.
  }
  \label{fig:supp_render}
  \Description{supp_render}
  \vspace{0.06cm}
\end{figure*}

\begin{figure*}[t]% specify a combination of t, b, p, or h for top, bottom, on its own page, or here
  \centering % avoid the use of \begin{center}...\end{center} and use \centering instead (more compact)
  \setlength{\abovecaptionskip}{2pt}  
  \setlength{\belowcaptionskip}{0pt}
  \includegraphics[width=0.85\linewidth]{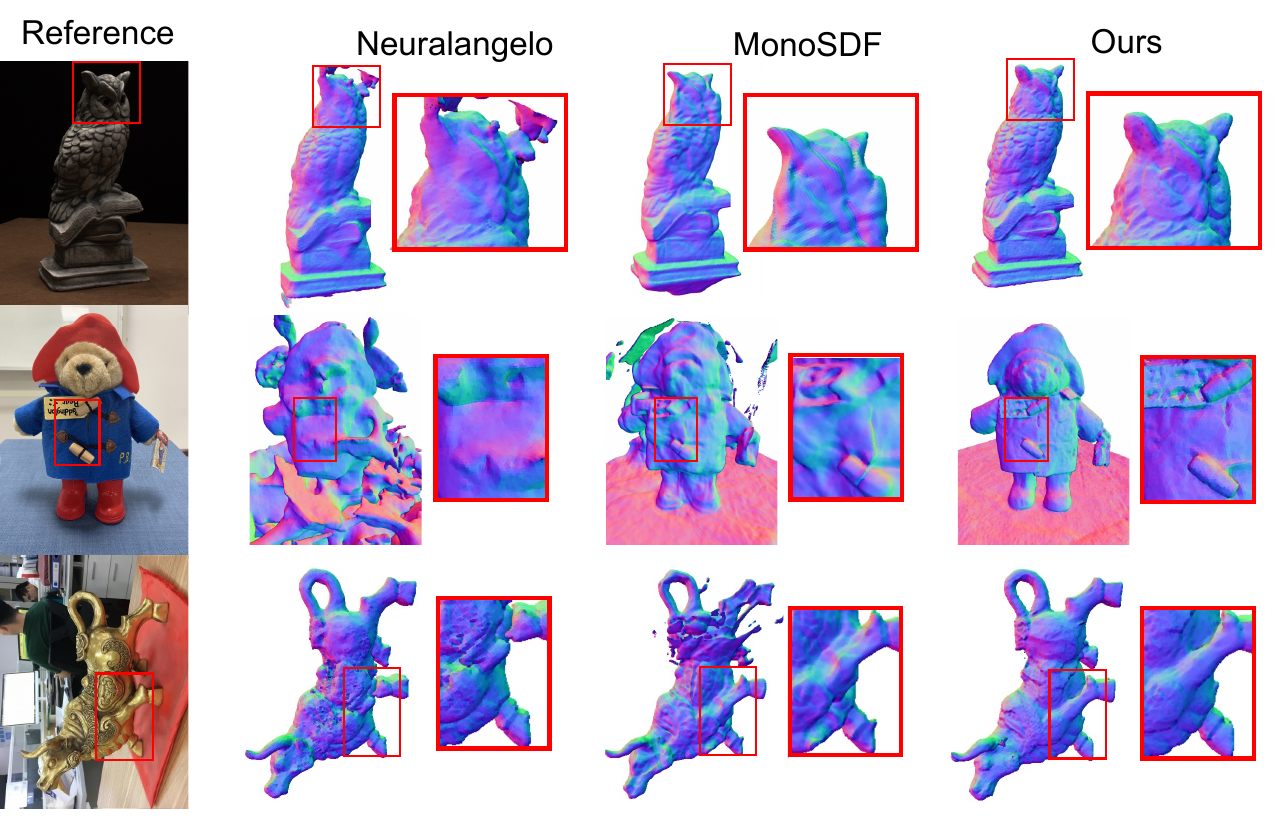}
  \caption{%
    Additional comparison results of surface normal on DTU and BlendedMVS. Note that our method produces more accurate and delicate results.
  }
  \label{fig:supp_normal}
  \Description{supp_normal}
  \vspace{0.03cm}
\end{figure*}